\pdfoutput=1

\documentclass[11pt]{article}

\usepackage{acl}

\usepackage{times}
\usepackage{latexsym}
\usepackage{amsmath}
\usepackage{bm}
\usepackage{amsfonts}
\usepackage{amssymb}
\usepackage{booktabs}
\usepackage{subcaption}
\usepackage{caption}
\usepackage{graphicx}
\usepackage{multirow}
\usepackage{arydshln}
\usepackage{soul, color}

\usepackage[T1]{fontenc}

\usepackage[utf8]{inputenc}

\usepackage{microtype}

\newcommand\blfootnote[1]{%
  \begingroup
  \renewcommand\thefootnote{}\footnote{#1}%
  \addtocounter{footnote}{-1}%
  \endgroup
}

\title{\includegraphics[width=0.05\textwidth]{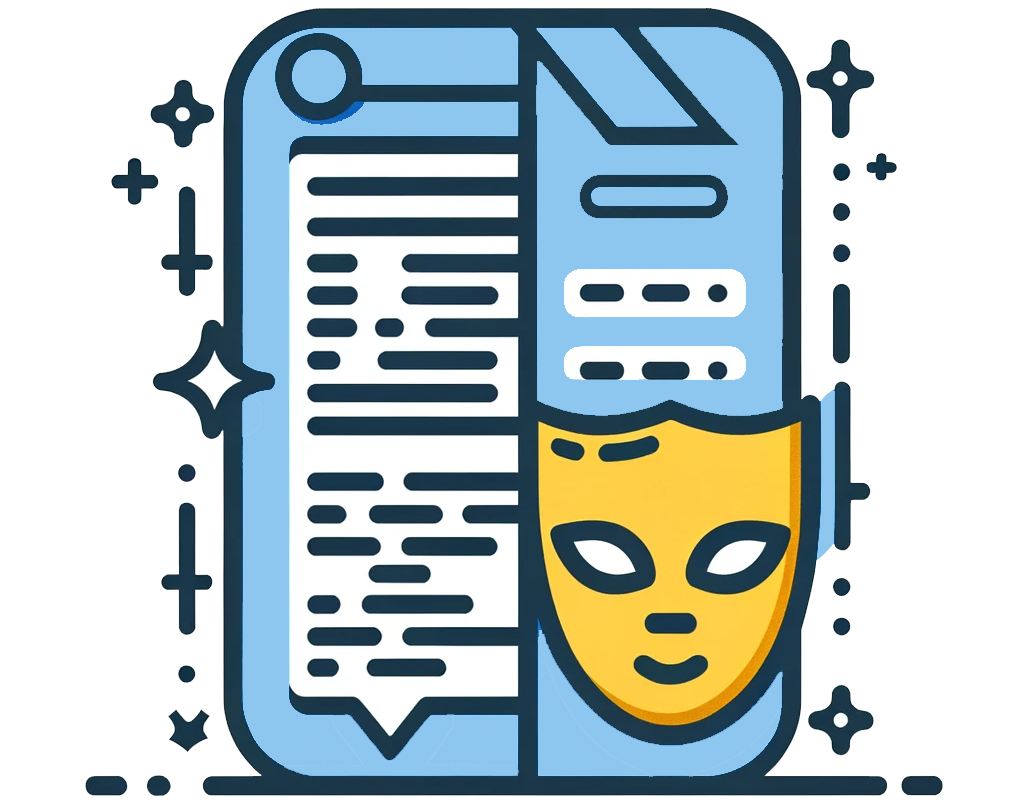}From Role-Play to Drama-Interaction: An LLM Solution}

\author{
Weiqi Wu$^\Diamond$\textsuperscript{\rm 1,2},
Hongqiu Wu$^\Diamond$\textsuperscript{\rm 1,2},
Lai Jiang\textsuperscript{\rm 1,2},
Xingyuan Liu\textsuperscript{\rm 1,2},
Jiale Hong\textsuperscript{\rm 1,2},\\
\textbf{Hai Zhao}$^\dag$\textsuperscript{\rm 1,2}
\and \textbf{Min Zhang}\textsuperscript{\rm 3} \\
\textsuperscript{\rm 1}Department of Computer Science and Engineering, Shanghai Jiao Tong University \\
\textsuperscript{\rm 2}Key Laboratory of Shanghai Education Commission for Intelligent Interaction \\
and Cognitive Engineering, Shanghai Jiao Tong University, Shanghai, China \\
\textsuperscript{\rm 3}Harbin Institute of Technology, Shenzhen, China \\
\texttt{\{wuwq1022,wuhongqiu\}@sjtu.edu.cn,}\\
\texttt{zhaohai@cs.sjtu.edu.cn,minzhang@suda.edu.cn}}

\begin{document}
\maketitle

\blfootnote{$^\Diamond$ Equal contribution.}
\blfootnote{$^\dag$ \,Corresponding author; This research was supported by the Joint Research Project of Yangtze River Delta Science and Technology Innovation Community (No. 2022CSJGG1400).}

\begin{abstract}
Drama is a form of storytelling inspired by human creativity, proceeding with a predefined storyline, carrying emotions and thoughts.
This paper introduces \emph{LLM-based interactive drama}, which endows traditional drama with an unprecedented immersion, where a person is allowed to walk into it and interact with the characters and scenes.
We define this new artistic genre by 6 essential elements—plot, character, thought, diction, spectacle and interaction—and study the entire pipeline to forge a backbone \emph{drama LLM} to drive the playing process, which is challenged by limited drama resources, uncontrollable narrative development, and complicated instruction following.
We propose \emph{Narrative Chain} to offer finer control over the narrative progression during interaction with players;
\emph{Auto-Drama} to synthesize drama scripts given arbitrary stories;
\emph{Sparse Instruction Tuning} to allow the model to follow sophisticated instructions.
We manually craft 3 scripts, \emph{Detective Conan}, \emph{Harry Potter}, \emph{Romeo and Juliet}, and design a 5-dimension principle to evaluate the drama LLM comprehensively.

\end{abstract}
\section{Introduction}
Large Language Models (LLMs) \citep{DBLP:journals/corr/abs-2303-08774,DBLP:journals/corr/abs-2307-09288,DBLP:journals/corr/abs-2310-06825,DBLP:journals/corr/abs-2309-10305, qwen} make powerful role-play systems. Their adeptness in natural language understanding and generation allows them to play diverse characters, based on user-specified descriptions and profiles \cite{Shanahan2023RolePW,DBLP:journals/corr/abs-2404-00276}.
However, our vision goes beyond mere role-play, as we attempt to empower LLMs to render the entire dramatic narrative rather than solely a single character.

\begin{figure*}[t]
    \centering
    \includegraphics[width=0.99\linewidth]{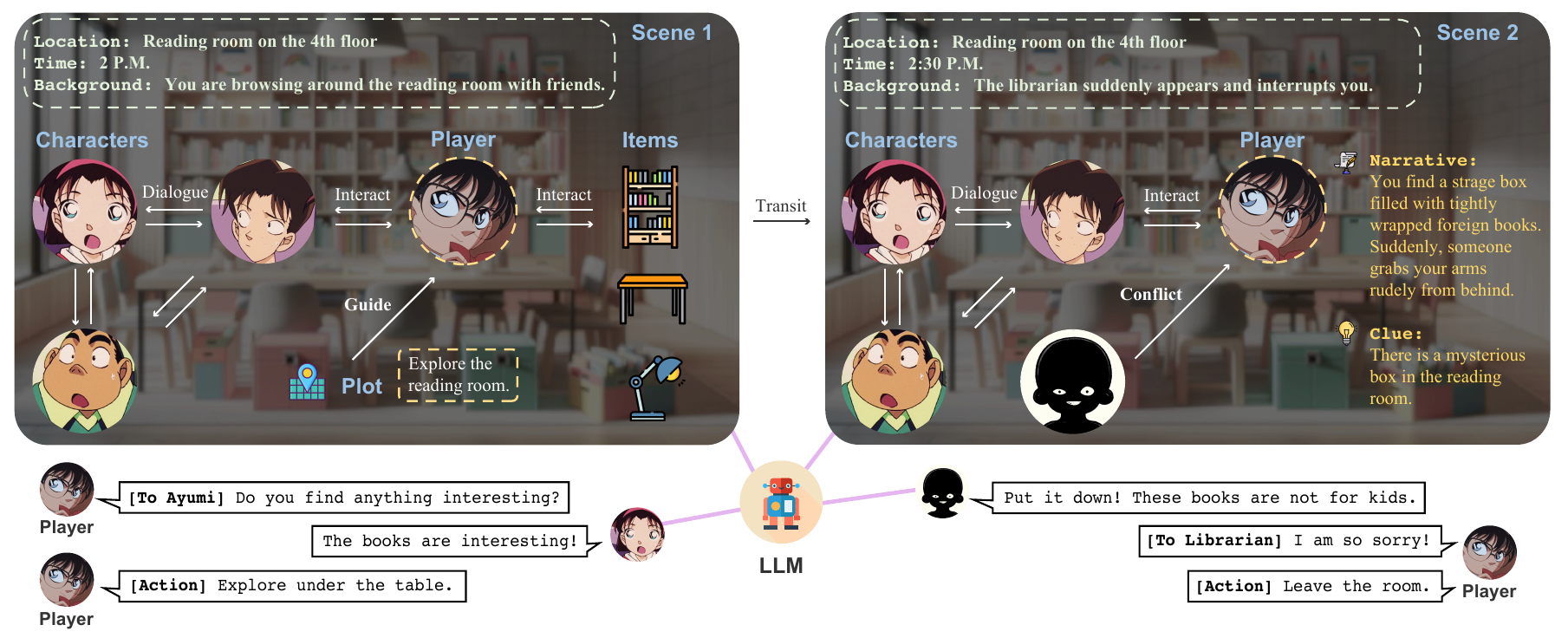}
    \caption{Illustration of LLM-based interactive drama. Two scenes from a script for \emph{Detective Conan (Library Murder Case, Episode 50)}, where the player takes the role of \emph{Conan} and there are other four characters, \emph{Ayumi}, \emph{Mitsuhiko}, \emph{Genta}, and \emph{Librarian}. The player can either make conversations with characters or take actions to progress the plot.}
    \label{fig:scene}
\end{figure*}

\emph{What is LLM-based interactive drama}

Drama is a classical mode of storytelling, based upon dialogues and performances with a multitude of scenes of a story.
In this paper, we study \emph{LLM-based interactive drama}, a new immersive mode of storytelling, where the audience can walk into the story and interact with the characters and environment in it.
Inspired by six essential elements of traditional drama outlined by Aristotle, i.e., plot, character, thought, diction, spectacle and music \cite{Laurel1991ComputersAT,betti2015introduction}, we outline the new six essential elements for interactive drama—plot, character, thought, diction, spectacle, and interaction. \textbf{Plot} presents a dramatic storyline through a series of scenes. \textbf{Character} refers to the roles participating in each scene, while \textbf{thought} delves into their inner motivations and psychological dynamics as the plot progresses. \textbf{Diction} refers to the dialogues among characters and the audience. \textbf{Spectacle} incorporates vivid text-based illustrations of the scene, such as background and items. Lastly, \textbf{interaction} stands as a pivotal and evolutionary element, fostering an immersive engagement between the audience and drama. Rather than role-play, the form of interaction in interactive drama is diverse and free. The audience is expected to have conversations with any character and take any action in the environment.

We show an example of interactive drama in Figure \ref{fig:scene}.
In Scene 1, for instance, there are three characters and three items, with a predefined narration as well as character dialogue for scene rendering.
The audience (as a player) is allowed the liberty to interact with any character (e.g., make a conversation to \emph{Ayumi}) and any item (e.g., explore under the table). The LLM serves as the characters and the items in the scene to respond to the player.
There is a target in the scene (``explore the reading room''). The purpose of it is to guide the player and all characters towards unfolding the plot smoothly.
As the storyline progresses, a scene transition will be triggered, bringing a new plot, e.g., new characters and dialogues, as shown in Scene 2.

\emph{How LLMs perform interactive drama}

We denote the LLMs that perform interactive drama as \emph{drama LLMs}.
Compared to playing a single character, there is an entire dramatic story inside the LLM, which means it is required to tackle the interplay of multiple characters and the audience, as well as progress the plot.
Therefore, it presents a more ambitious challenge to LLMs, extending their adeptness beyond the mere simulation of characters.

In contrast to playing a role instructed by its profile, the drama LLM follows a more intricate instruction.
In this paper, we present a prototype \textbf{drama script}, a global instruction to guide the drama LLM to orchestrate the drama.
A drama script is not a single instruction but a collection of individual scenes. In each scene, we detail the spectacle, characters, plot, etc. To assist the drama LLM in better managing the interplay between player agency and scripted plot progression, we introduce the novel concept of \textit{Narrative Chain}, which divides the narrative into smaller consecutive segments. By guiding the player through each segment through the interaction with drama LLM, the player is allowed to explore and influence the plot autonomously while experiencing a coherent and smooth story progression.

\emph{How one trains and evaluates drama LLMs}

We propose a comprehensive training paradigm to fine-tune a general drama LLM to decently play the drama from given scripts, rather than zero-shot prompting.
There are two main challenges.
First, it is an exhausting process to acquire a large number of drama scripts.
We propose \emph{Auto-Drama}, an efficient data pipeline to generate the drama scripts automatically from arbitrary stories based on GPT-3.5 \citep{DBLP:journals/corr/abs-2303-08774}, including scene extraction, plot production, and trigger imagination.
Moreover, a drama script is a lengthy and sophisticated instruction encompassing a series of sub-tasks and only a small fraction of them will be activated during each inference, which is based on the player's behaviour.
It incurs a hard learning process for the LLM to follow such instructions.
We propose \emph{Sparse Instruction Tuning (SIT)}, a two-stage training that unlocks more accurate instruction following.
Eventually, we propose a fine-grained evaluation process, measuring the performance from 5 dimensions - scenery, narration, coherency, guidance, and transition.
A series of cases are illustrated to further demonstrate the ability of drama LLMs.

In summary, this paper:

$\bullet$ introduces an LLM-based solution for interactive drama with a prototype drama script;

$\bullet$ proposes the data generation technique that fuels the learning of drama LLMs;

$\bullet$ proposes an enhanced instruction tuning technique to train drama LLMs;

$\bullet$ presents the multi-aspect evaluation to assess the performance of drama LLMs.

\section{Related Works}

\paragraph {Role-play LLMs}
Simulating specific characters in conversation is a popular topic in Artificial Intelligence (AI) research \cite{zhang2009drama, avin2020exploring}. Recently, LLMs have been used to mimic characters with various attributes and conversational styles \cite{Shanahan2023RolePW, wang2023does, tu2024charactereval}. There are generally two approaches to building role-play LLMs. The first involves prompting the model with detailed character profiles \cite{li2023chatharuhi, tao2023rolecraftglm, Wang2023RoleLLMBE, chen-etal-2023-large} or specific utterances \cite{han-etal-2022-meet}. The other involves fine-tuning the model on a character's experiences and dialogues \cite{shao-etal-2023-character, Zhou2023CharacterGLMCC, lu2024large}. The success of role-play LLMs lays the basis for playing characters in drama to create more immersive experiences.

\paragraph{Interactive Narrative Intelligence} 
Interactive narrative \citep{bates1991broad} is a dynamic form of digital interactive experience where users can affect the storyline \cite{mateas2000neo, szilas2007computational, riedl2013interactive}. It can be applied in both entertainment \cite{riedl2012interactive, yong2023playing, DBLP:journals/corr/abs-2404-00276} and serious domains like education \cite{plowman2014getting, wang2017interactive}. LLMs have been explored for tasks such as crafting scenes \cite{Kumaran2023SceneCraft} and shaping pivotal plot points \cite{harmon2023prompt}, further pushing the boundary of interactive narratives. \citet{zhao2023narrativeplay} interacts with fictional characters in narratives in a multi-modal environment with LLMs. We aim to interact with not only the characters but also the plot and scenes, constructing a more immersive interactive drama based on LLMs.

\paragraph{AI for Art}
AI has revolutionized art creation, with applications ranging from generating paintings \cite{Castellano2021DeepLA}, music \cite{hernandezolivan2022survey, zhu2023survey}, poems \cite{chakrabarty-etal-2022-help, bena-kalita-2019-introducing} to crafting screenplays and theatre scripts \cite{10.1145/3544548.3581225, Pramnik2022SurveyAM}. Additionally, AI enhances artistic expression through interactive experiences \cite{digital2010002, zhao2023narrativeplay}. In this paper, we explore the augmentation of artistic expression through user-LLM interaction by constructing drama LLMs. 

\section{Interactive Drama}

In this section, we redefine elements of traditional drama from Aristotle's dramatic theory to propose the new six elements for \emph{LLM-based interactive drama}.
Upon the new definition, we then present the prototype of a drama script, by which LLMs are instructed to perform the drama.

\begin{figure*}[h]
    \centering
    \includegraphics[width=0.99\linewidth]{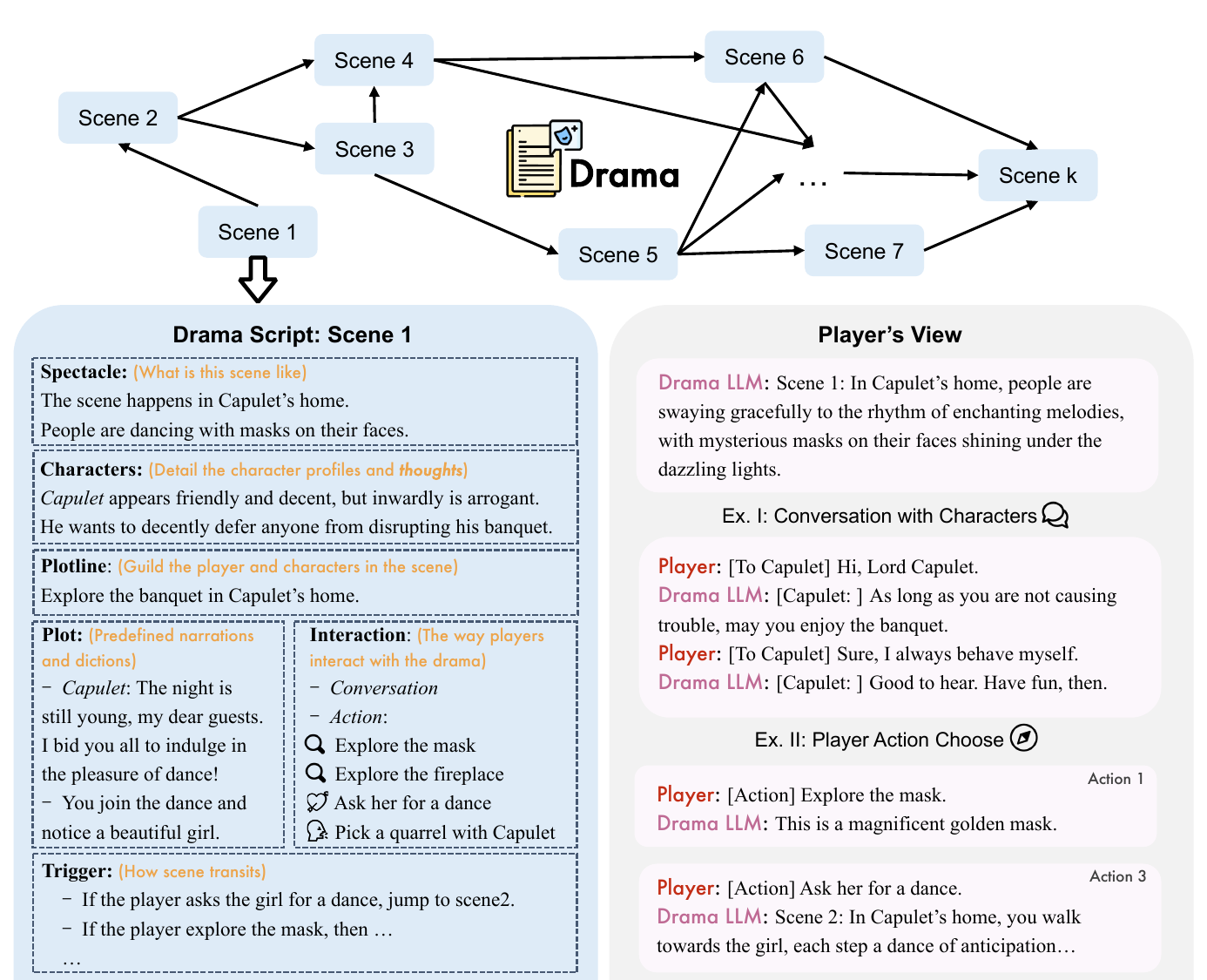}
    \caption{Prototype of drama scripts to prompt drama LLMs. The identifier at the beginning of each input either designates a character for dialogue or indicates that the input corresponds to an action outlined in the script.}
    \label{fig:drama-play}
\end{figure*}

\subsection{LLM-based Interactive Drama}

AI-based interactive drama has been mentioned by creative researchers for many decades \citep{laurel1986toward,DBLP:journals/presence/Bates92,mateas2000neo}.
We focus on LLM-based interactive drama in this paper, which comprises six essential elements, i.e., plot, character, thought, diction, spectacle, and interaction. They work in synergy to craft an immersive and dynamic storytelling experience.

\paragraph{Plot} Plot is the backbone of the storytelling. Generally, a drama tells the story with a sequence of scenes. The transition of scenes suggests the development of the plot. In contrast to the fixed plot in traditional drama, interactive drama allows the audience to influence the development of the story through interactions, to some extent. Therefore, it is crucial for drama LLMs to process the interplay between the audience and the plot.

\paragraph{Character} Characters are the individuals who inhabit the dramatic world in each scene, defined by various settings like personalities, motivations and behaviours. Rather than playing a single character in traditional role-play, drama LLMs play multiple characters in the drama and process the relationships between them simultaneously.

\paragraph{Thought} Thought is an important part of the character settings, representing their inner motivation for behaviours. Thoughts of characters may change as the plot develops. Hence, it is necessary to update the character settings provided for LLMs when scenes switch to ensure a stable memory. 

\paragraph{Diction} Diction in traditional drama refers to the preset dialogues between characters. Interactive drama enables dynamic conversations between characters and the audience.
In both situations, the characters should be coherent with their settings.

\paragraph{Spectacle} While traditional drama relies on visual assets to create a spectacle, LLM-based interactive drama utilizes text to display. A descriptive tone is leveraged to recover, refine and render the scenery including background and items, thereby immersing the audience in the dramatic world.

\paragraph{Interaction} Interaction emerges as a new element for interactive drama. It bridges the audience with dramatic storytelling, where the audience turns into the player and engages in the story. Generally, the player is allowed to converse with characters or perform specific actions within the scene. These interactions may impact the plot by triggering new events or changing the characters' endings.

\subsection{Drama Script}
Upon interpreting six basic elements describing LLM-based interactive drama, we materialize them into a new style of instruction for LLMs to perform the interactive drama, named \textbf{drama script}.

A drama script serves as the global instruction, outlining the desired drama for drama LLMs. In the prototype presented in Figure \ref{fig:drama-play}, we demonstrate how the scripts can be tailored to specific narratives, allowing for creative adaptation.
Overall, a drama script is a network of individual scenes that collectively form the storyline, which can unfold linearly or non-linearly. Each scene transitions to the next based on specific conditions being met.
Specifically, every single scene is a sophisticated instruction detailed with six parts, as shown in the left side of Figure \ref{fig:drama-play}.
\textbf{Spectacle} describes the scenery. Once transiting to a new scene, the model will be triggered to render it to the audience.
What follows is the character settings in \textbf{Character}. The player can have a conversation with any of the characters in the scene.
Notably, in contrast to open-ended playgrounds, interactive drama should follow an underlying storyline. To balance player autonomy and the smooth unfolding of the plot, the main target of the current scene is stated in \textbf{Plotline}, which guides the characters to avoid unrelated or even offensive discussions with the audience and pull them back to the main plot.
\textbf{Plot} lists the preset dialogue among characters and narration in the scene.
\textbf{Interaction} defines the way that the player can interact. In this prototype, we offer two main forms of interaction, i.e. having conversations and choosing specific actions. \textbf{Trigger} defines how the player's behaviour impacts the plot, e.g., scene transition, discovering new clues, and having a new relationship with some character. The drama LLM should learn to be a perfect trigger to unfold the plot.

Each model input includes scripts of relevant scenes and multi-turn user inputs. As depicted in the right half of Figure \ref{fig:drama-play}, the drama LLM initially generates a description of the spectacle and plot, as well as renders the available interactions to the player. Subsequent user inputs take the form of either dialogue with characters in the scene, specified by the character's name followed by the dialogue content, or actions, composed of an action identifier and the chosen action. The drama LLM then processes these inputs, either role-playing characters to respond to the user or producing corresponding plot updates based on the trigger.

\section{Narrative Chain}

Interactive drama presents a critical challenge in balancing player agency with the story authored \citep{weyhrauch1997guiding,magerko2005story}. While players influence the narrative progression through dialogues and actions, there remains a need to guide them to explore the pre-drafted story, which should be done by the drama LLM.

\begin{figure}[ht]
    \centering
    \includegraphics[width=0.99\linewidth]{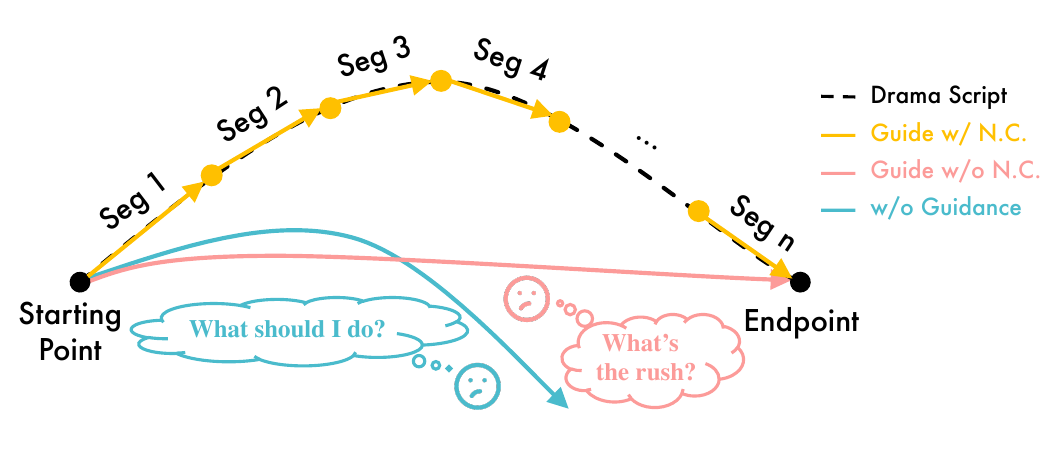}
    \caption{Story arcs defined by the drama script and experienced by the player across different methodologies.}
    \label{fig:nc}
\end{figure}

Any narrative has a beginning and an end, which can be visualized as two points in the space, as depicted in Figure \ref{fig:nc}. Maintaining its progression requires the player to reach the endpoint from the starting point. Otherwise, the player may get lost and the subsequent story cannot progress properly (the blue line). A straightforward way is to intervene in the player's behaviours through interaction. For instance, the drama LLM repeatedly asks the player to take specific actions to ensure that the narrative reaches its intended endpoint. However, this method can be obtrusive and greatly damages the player experience (the pink line). Therefore, we propose a novel concept termed \textit{Narrative Chain}, which defines the way the drama LLM guides the story progression smoothly and coherently.

A drama unfolds through a story arc, which is the trajectory formed by linking all the narratives together. In a specific scene, this arc can be divided into smaller narrative segments, resembling a chain of sub-narratives. By navigating the player through each segment, we can establish a nearly straight line between every two adjacent points, approximating the curvature of the overall story arc, as illustrated in Figure \ref{fig:nc}. This segmentation facilitates finer control over the narrative development, ensuring a smooth progression throughout the story arc. Ultimately, the step-by-step guidance directs the player through a sequence of narratives while respecting interactive freedom, as the behaviour of the player remains open-ended.

Therefore, when the narrative of a scene is complex, it can be effectively presented in the form of a chain. For example, in Scene 1 of Figure \ref{fig:scene}, to guide the player to explore the reading room, the plotline ``Explore the reading room'' can be further divided into ``Search for books with Genta - Ayumi wants to do homework by the table - Ayumi discovers a box under the table''. Players gradually delve deeper into the narrative, gaining a deeper understanding of the unfolding drama before they encounter the core tasks predefined by the script. Meanwhile, the drama LLM is tasked with assessing which stage of the Narrative Chain it is currently navigating and providing accurate guidance relevant to that specific sub-narrative to ensure that each segment is thoroughly explored and complete. Additionally, the drama LLM determines when to introduce the next segment, facilitating the completion of the narrative development within the scene.

\section{Data Generation}

\begin{figure*}[ht]
    \centering
    \includegraphics[width=\linewidth]{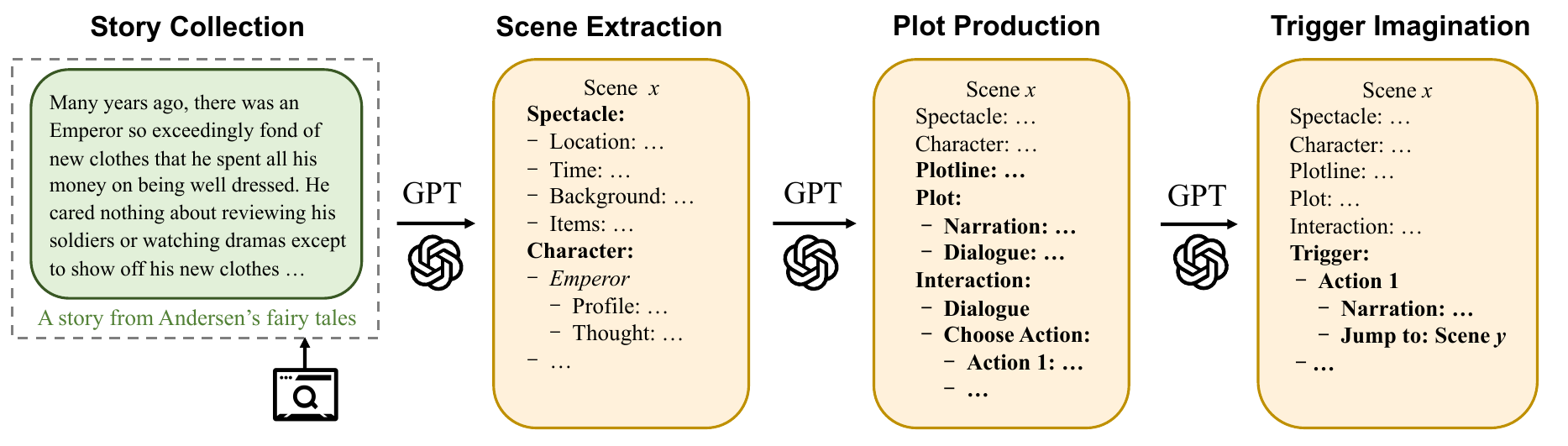}
    \caption{The pipeline of Auto-Drama for drama script generation.}
    \label{fig:data}
\end{figure*}

Harnessing LLMs to construct interactive drama necessitates the process of fine-tuning on a large amount of supervised data, while manually creating diverse drama scripts is a tough process.
It will be nice and efficient to generate drama scripts automatically from public stories on the web.
We thus propose \textit{Auto-Drama} that crafts drama scripts from any given story harnessing the power of GPT-3.5.
Rather than prompting GPT-3.5 to directly draft the script from the story, incurring low-quality and incomplete drama,
\textit{Auto-Drama} is a pipeline featured by four main steps, as illustrated in Figure \ref{fig:data}. The process enriches a brief scene to a detailed one while ensuring it fulfils the six elements of interactive drama.

\paragraph{Story Collection}
First, we crawl Andersen's Fairy Tales from the web and collect over 80 stories.
This corpus serves as the foundational resource for our subsequent script generation phases.

\paragraph{Scene Extraction} 
Given a story, we prompt GPT-3.5 to imagine itself as one of the main characters in the story and break down the story into a set of scenes from the view of a player. Simultaneously, we task it with generating scene details. 
We request basic information about the scene which helps drama LLMs to recover it, including location, time, atmosphere and items. We also query GPT-3.5 to enrich each character with details regarding their personalities, thoughts and behaviours.

\paragraph{Plot Production} 
\label{sec:flow}
Given the scene, the next significant step is crafting the plot within it, requiring associated story content. 
The plot includes three aspects.
The first is the plotline to ensure a smooth and uninterrupted progression of the story. 
The second is the performance in the scene, which is a detailed plot consisting of two forms of performance: dialogue among characters and background narration.
The last is the potential interactions for the player, involving conversation and action.
We prompt GPT-3.5 to determine if the player can engage in conversation within the scene, excluding scenarios where urgency precludes chat or no other characters are present. 
In addition, we prompt it to imagine the possible actions for the player related to the plot, by offering it some examples.

\paragraph{Trigger Imagination}
Lastly, we prompt GPT-3.5 to imagine the possible consequences triggered by the player interaction.
For example, some words said and some choices made by the player may lead to changes in the characters' behaviour. These changes can impact the progress of the story, which cannot be gained by simply playing the characters based on their settings.
We also request at least one transitional consequence to connect individual scenes seamlessly.

\paragraph{Dialogue and Narrative Generation for Auto-Drama}
We acquire sufficient drama scripts thanks to Auto-Drama. One step left is to acquire the training samples (i.e. input and output contents) for the script, which are possible interactions between the LLM and the player.
Since the player's actions are provided as options, we focus on generating dialogue and narrative content with the assistance of GPT-3.5.
We instruct GPT-3.5 to create dialogue based on triggering conditions for the player's proactive speech. 
We additionally generate casual conversations between all characters and the player, following the character settings.
For LLMs to guide the dialogue, we provide examples of unrelated dialogues that require guidance and prompt GPT-3.5 to generate more. 
The number of rounds is limited to 2 to 5. Ultimately, we produce refined scene descriptions and narrative dialogues for the plot. Paraphrasing and rephrasing prompts are employed to ensure the diversity of data.

\section{Sparse Instruction Tuning}

In this section, we present the detailed methodology to train drama LLMs. Based on our prototype drama script, drama LLMs will learn to follow a lengthy and sophisticated instruction that encompasses a series of sub-tasks. For instance, on top of general language understanding and generation, a drama script covers sub-tasks like:
$\bullet$ \textit{Transition}: locate the next scene;
$\bullet$ \textit{Refinement}: refine scenery given the narrative tone;
$\bullet$ \textit{Role-Play}: play characters following their profiles and thoughts; 
$\bullet$ \textit{Guidance}: guide the player back to the plotline;
$\bullet$ \textit{Semantics}: capture the nuanced semantics of the player's talk to trigger the correct consequence;
etc.

\begin{figure}[t]
    \centering
    \includegraphics[width=0.99\linewidth]{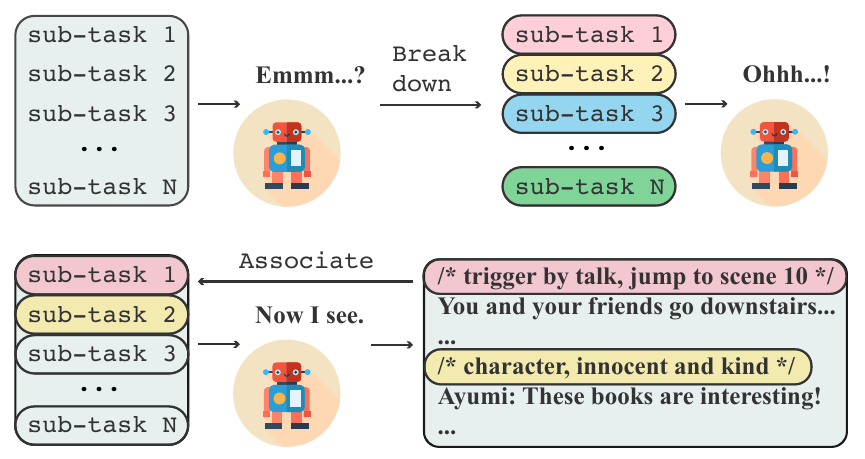}
    \caption{Diagram of Sparse Instruction Tuning.}
    \label{fig:sit}
\end{figure}

\begin{figure*}[t]
    \centering
    \includegraphics[width=0.9\linewidth]{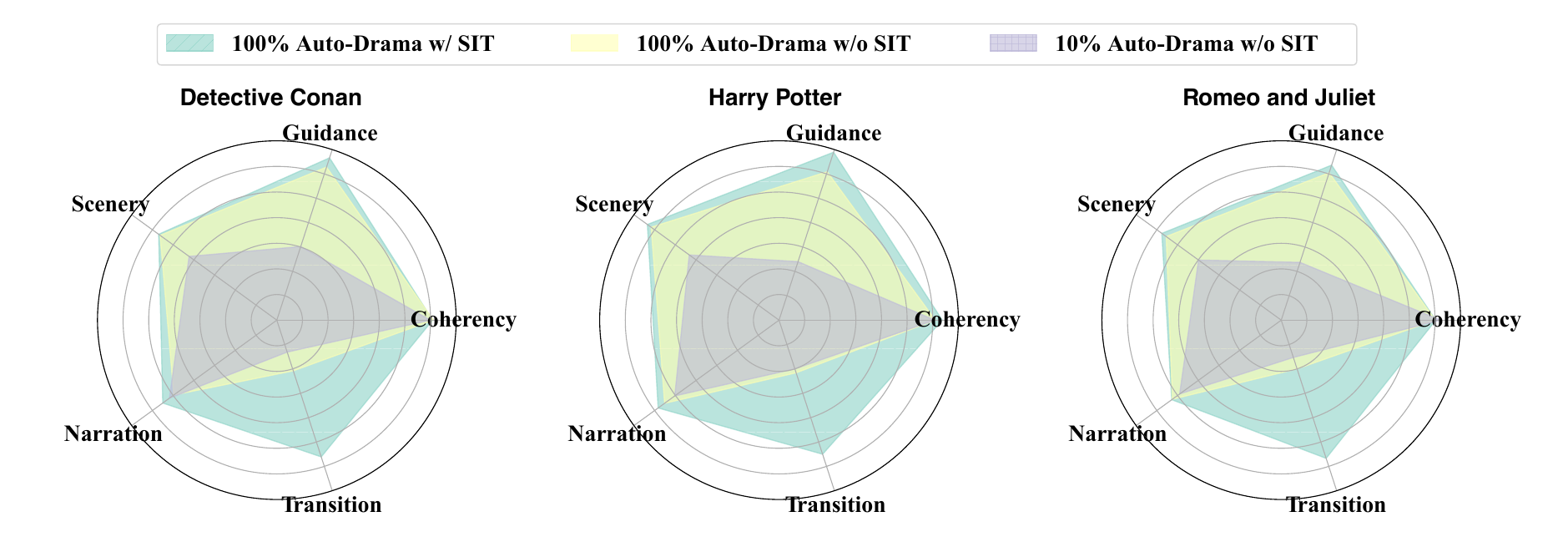}
    \caption{Performance of 8B drama LLMs across our three drama scripts with different training setups: (1) fine-tuned with completed Auto-Drama data and Sparse Instruction Tuning (SIT), (2) fine-tuned with completed Auto-Drama data without SIT, (3) fine-tuned with only 10 \% proportion of Auto-Drama data without SIT.}
    \label{fig:exp}
\end{figure*}

However, it is challenging for the model to navigate all sub-tasks within one instruction at the same time.
Furthermore, for each inference, only a small proportion of sub-tasks are associated.
We denote such instructions as \emph{sparse instructions}, which make the fine-tuning process pretty hard and in low efficiency.
To this end, we propose \emph{sparse instruction tuning (SIT)}, which contains two stage as shown in Figure \ref{fig:sit}: (1)
In the first stage, we break down the drama script into a series of sub-tasks and craft an individual instruction for each sub-task to describe it.
As a result, the fine-tuning process starts with training the model on all these sub-task instructions, which offers a nice initial state for the subsequent learning \cite{DBLP:conf/emnlp/Wu0ZZ23}. 
(2) In the second stage, where the model undergoes fine-tuning on entire drama scripts, we require the model to refer to the associated sub-tasks in responses. 
Specifically, we introduce an annotation ``/*'' and ``*/''  to encapsulate the sub-tasks which prefixes the associated contents in the response, which can be filtered out with ease when presented to the audience.
This trick effectively creates a link between the instruction and sparse sub-tasks, thus allowing for more accurate instruction following.

\section{Experiments}

We fine-tune LLaMA3-8B-Instruct\footnote{https://huggingface.co/meta-llama/Meta-Llama-3-8B-Instruct} \cite{DBLP:journals/corr/abs-2307-09288} and Qwen1.5-14B-Chat\footnote{https://huggingface.co/Qwen/Qwen1.5-14B-Chat} \cite{qwen} on data generated by Auto-Drama (\textit{Auto-Drama data} for simplicity). Detailed training setups are in Appendix \ref{app:setup}. We evaluate drama LLMs on three manually-written scripts: a detective story (adapted from \textit{Detective Conan}), an adventure story (adapted from \textit{Harry Potter}) and a classical drama (adapted from \textit{Romeo and Juliet}).

\subsection{Evaluation Setup}

From a plot-centric perspective, we propose five critical dimensions for assessing the efficacy of drama LLMs:

\paragraph{Scenery} This dimension evaluates the scene presentation by drama LLM, considering how well it aligns with the provided details and intended tone.

\paragraph{Narration} Similar to scenery but focusing on a different aspect (plot v.s. spectacle), it assesses how effectively the plot narration aligns with the intended tone and atmosphere of the scene. 

\paragraph{Transition} We examine the effectiveness of drama LLMs in managing scene transitions, ensuring that the scene changes appropriately when triggered by the player.

\paragraph{Guidance} We assess how decent drama LLMs maintain the player engagement with the plotline during the interaction, ensuring players stay connected to the plot and smoothly unfold the plot.

\paragraph{Coherency} This dimension evaluates the adeptness of drama LLMs in representing characters, and whether responses by characters align with their established profiles and internal thought processes.

GPT-4 is employed as the judge to score scenery, narration, guidance, and coherency on a 7-point Likert scale. We manually check the transition score to accurately examine whether the drama LLM transits to a new scene or stays in the current scene correctly. We assign a score of 7 points for a correct transition, 4 points if the trigger annotation is correct but the transition is wrong, and 1 point for any other situations.

\subsection{Results and Discussion}

\paragraph{Overall Results}

Figure \ref{fig:exp} illustrates the performance of our 8B drama LLM. Trained on Auto-Drama data with sparse instruction fine-tuning, the drama LLM achieves exceptional scores across all dimensions. It demonstrates remarkable capabilities in engaging in dialogue with players, generating fluent and rich narratives based on the plot, and accurately handling plot progression. Notably, the model excels in guidance, effectively steering the player back to the main storyline through generated responses and ensuring correct plot progression amidst various transitions.

However, as elaborated in Appendix \ref{app:14b}, scaling up to the 14B model does not yield substantial improvements except for transition, which is a relative weakness for smaller drama LLMs. This could suggest that larger models hold an advantage in handling more complex scenarios, while increasing model size may not always be beneficial for certain tasks, especially if the task itself is not complex enough to warrant the additional model capacity.

\paragraph{Effectiveness of Auto-Drama}

To assess the impact of Auto-Drama data, we conduct experiments by fine-tuning LLaMA3-8B-Instruct with varying proportions of Auto-Drama data. From Figure \ref{fig:exp}, comparing the drama LLM trained on 100\% and 10\% of Auto-Drama data, we observed a significant improvement in scenery and guidance. Despite the limited amount of training data, drama LLM trained on 10\% of the Auto-Drama exhibits notable performance on coherency and narration due to its inherent zero-shot role-playing and text integration capabilities. The richer dataset further enhances its ability to depict scenes and guide dialogues effectively, validating the effectiveness of Auto-Drama.

\paragraph{Effectiveness of SIT}

Despite the promising results obtained from training the drama LLM using Auto-Drama, its accuracy remains low in complex sub-tasks within the instruction, such as transition, where it needs to identify the player's intent in the dialogue. Figure \ref{fig:exp} shows that SIT effectively enhances the ability to follow complex instructions, resulting in improvements across all dimensions, especially transition. Therefore, SIT proves to be a highly effective training method for complex instruction following, which is significant for performing interactive drama.

\section{Case Study}

\begin{table*}[ht]
    \scriptsize
    \centering
    \resizebox{0.99\linewidth}{!}{
    \begin{tabular}{@{}p{0.43\linewidth}p{0.43\linewidth}@{}}
    \toprule
    \textbf{Case 1: Narrative Tone} \\
    \midrule
    
   \textcolor[HTML]{000000}{\textit{Tone: Mysterious and Tense}}
    
     
    \textcolor[HTML]{DB7093}{\textbf{Drama LLM:}} Hermione \hl{whispered}, ``What's that by the three-headed dog?'' Ron replied \hl{cautiously}, ``It looks like a harp.''

    &

    \textcolor[HTML]{000000}{\textit{Tone: Relaxed and Cheerful}}
    
     
    \textcolor[HTML]{DB7093}{\textbf{Drama LLM:}} Hermione asked \hl{with a smile}, ``What's that by the three-headed dog?'' Ron \hl{excitedly} replied, ``It looks like a harp.''
    
    \\
    \midrule
    \midrule
    \textbf{Case 2: Scene Transition}\\
    \midrule

    \textcolor[HTML]{000000}{\textit{Trigger 1: the player thinks the books were still in the room}}
    
    \textcolor[HTML]{458B74}{\textbf{Player: }}To Ayumi: There are so many empty shells that \hl{there is no way to fit them all into the purse}.

    \textcolor[HTML]{DB7093}{\textbf{Drama LLM:}} You realize that some of the books must be hidden somewhere in the room. Therefore, you and your friends decide to search for them. \hl{(Jump to Scene 10)}
     
    &

    \textcolor[HTML]{000000}{\textit{Trigger 2: the player thinks the books were taken away}}
    
    \textcolor[HTML]{458B74}{\textbf{Player: }}To Ayumi: He should have \hl{taken all the books with him so as not to be discovered}. I guess we'll just have to find him.

    \textcolor[HTML]{DB7093}{\textbf{Drama LLM: }} You all think the books were taken away, so you decide to head downstairs to follow the suspect. However, you get caught by the suspect. \hl{(Jump to Scene 11)}
    \\
    \midrule
    \midrule
    \textbf{Case 3: Plot Guidance} \\
    \midrule

    \textcolor[HTML]{000000}{\textit{Plotline: Help Peter read the invitation in his hand.}}
    
    \textcolor[HTML]{458B74}{\textbf{Player: }}To Peter: What is your favourite color?

    \textcolor[HTML]{DB7093}{\textbf{Drama LLM: }}Peter: It's red. \hl{Sir, please focus and read the words on this invitation.}
     
    &

    \textcolor[HTML]{000000}{\textit{Plotline: null}}
    
    \textcolor[HTML]{458B74}{\textbf{Player: }}To Peter: What is your favourite color?

    \textcolor[HTML]{DB7093}{\textbf{Drama LLM: }}Peter: My favourite color is red.
    \\
    \midrule
    \midrule
    \textbf{Case 4 Role-Play} \\
    \midrule
    \textcolor[HTML]{000000}{\textit{Voldemort: Evil, ruthless, cruel and hateful. Will stop at nothing to obtain the Philosopher's Stone for resurrection.}}
    
    \textcolor[HTML]{458B74}{\textbf{Player: }}
    To Voldemort: I won't give you the Philosopher's Stone!
    
    \textcolor[HTML]{DB7093}{\textbf{Drama LLM: }}
    Voldemort: Humph, your resistance is futile. \hl{The Philosopher's Stone will be mine.}
     
    &

    \textcolor[HTML]{000000}{\textit{Voldemort: Kind, warm, friendly and polite. Wish to obtain the Philosopher's Stone and use it for good. }}

    \textcolor[HTML]{458B74}{\textbf{Player: }}
    To Voldemort: I won't give you the Philosopher's Stone!

    \textcolor[HTML]{DB7093}{\textbf{Drama LLM:}}
    Voldemort: Harry, you should know that opposing me will lead to no favourable outcome. However, you can choose to accept, and \hl{I will offer you certain privileges.}
    \\
    \bottomrule
    \end{tabular}
    }
    \caption{Case study of interactions between the player and drama LLM based on LLaMA3-8B-Instruct.}
    \label{tab:case-study}
\end{table*}

Table \ref{tab:case-study} takes a closer look at how drama LLMs work.

\paragraph{Narrative Tone} Case 1 showcases how different tones of a scene affect the narration generated by the drama LLM. When a specific tone is appointed, we can find the adeptness in modifying the narration to either cultivate an atmosphere of mystery and tension where characters engage in low-toned and cautious talks or to foster a relaxing and cheerful environment where characters exhibit smiles and excitement. This augments the narrative diversity of drama LLMs, better utilizing simplistic scripts to present varied and vivid dramatic content.

\paragraph{Scene Transition} Case 2 exemplifies the effectiveness of drama LLMs in managing scene transitions based on the player's input. By detecting the belief of the player regarding the presence or absence of books in the room, the drama LLM guides the narrative towards the appropriate scene, as dictated by triggers defined within the drama scripts. 
Triggering by dialogue requires an understanding of nuanced semantics within the input to match those defined triggers within the script. Misinterpretation of the intention may lead to erroneous and bad plot progression.

\paragraph{Guidance} In Case 3, given a plotline, the drama LLM showcases its adeptness in steering the focus of the player back to the storyline, particularly when responding to the player's inquiries that veer off-topic. Hence, drama LLMs should possess the capability to gently guide the audience back to the drama. This characteristic underscores the plot-centric nature of interactive drama.

\paragraph{Character Setting} Case 4 highlights the effectiveness of portraying diverse personalities. Provided with diametrically opposed personalities (evil v.s. kind) for the same character \emph{Voldemort}, the drama LLM exhibits corresponding variations in its responses to the same input. 

\section{Conclusion}

This paper introduces LLM-based interactive drama and proposes the paradigm to train drama LLMs to realize this innovative form of storytelling. Based on elements of interactive drama, a prototype of drama script is proposed to serve as the global instruction for drama LLMs. To offer finer control over the drama development without harming the interactive freedom of the audience, we segment the narrative to construct the narrative chain and have drama LLMs to navigate the audience through each sub-narrative. To facilitate the training process, Auto-Drama is proposed to automate script generation, as well as Sparse Instruction Tuning to help LLMs follow complicated instructions describing many sub-tasks. Through comprehensive evaluation, we show the performance of the trained drama LLMs.

\section*{Limitations}
While our drama LLMs offer exciting potential for immersive interactions, several limitations warrant further exploration: (1) Limited Modalities: Our current drama LLMs primarily support text-based interactions. Additional modalities such as images, sound, or video could enrich the immersive experience, but this expansion presents technical and design challenges. (2) Action Complexity: There are significant constraints on player-scene action interaction in our current setup. One possible avenue for exploration is the integration of physical models \cite{Hao2022PhysicsInformedML, Asri2022ResidualMR, Seyyedi2023MachineLA} to simulate more immersive physical interactions between players and scenes. (3) Evaluation: Despite our five-dimension automatic evaluation performed by GPT-4, a more robust assessing method is crucial for advancing LLM-based interactive drama. A large-scale survey among users could be valuable for gathering insights in future work.
These limitations highlight the importance of ongoing research and development efforts aimed at addressing the challenges associated with LLM-based interactive drama.

\section*{Ethics Statement}

The development and use of drama LLMs are guided by ethical principles to ensure responsible and beneficial outcomes. (1) Data: We utilize stories from Anderson's fairy tales for generating data, and make adaptations of \textit{Detective Conan (Library Murder Case)}, \textit{Harry Potter and the Philosopher's Stone}, and \textit{Romeo and Juliet} to construct the test set. To address potential ethical concerns related to the utilization of Narrative Chainyrighted materials, we affirm that our research is conducted for academic and non-commercial purposes only. The use of these texts is solely for the development and evaluation of our models in natural language processing tasks, aimed at advancing scientific knowledge in the field. (2) Responsible Usage: We encourage the responsible use of drama LLMs for educational, entertainment, and creative purposes while discouraging any harmful or malicious activities.

\bibliography{custom}

\appendix
\section{Supervised Fine-tuning Setup}
\label{app:setup}

During supervised fine-tuning, both models are trained for 5 epochs, using AdamW optimizer with a learning rate of a weight decay of 0.01 and a cosine learning rate decay for learning rate warmup. The learning rate is set to 3e-4 for the 8B model and 1.5e-4 for the 14B model. We adopt LoRA for more efficient training and set the LoRA rank to 8, alpha to 32 and dropout to 0.1. It takes around 5 hours to train an 8B drama LLM and 10 hours to train a 14B drama LLM.

\section{Model Scaling}

\begin{figure}[htbp]
    \centering
    \includegraphics[width=0.7\linewidth]{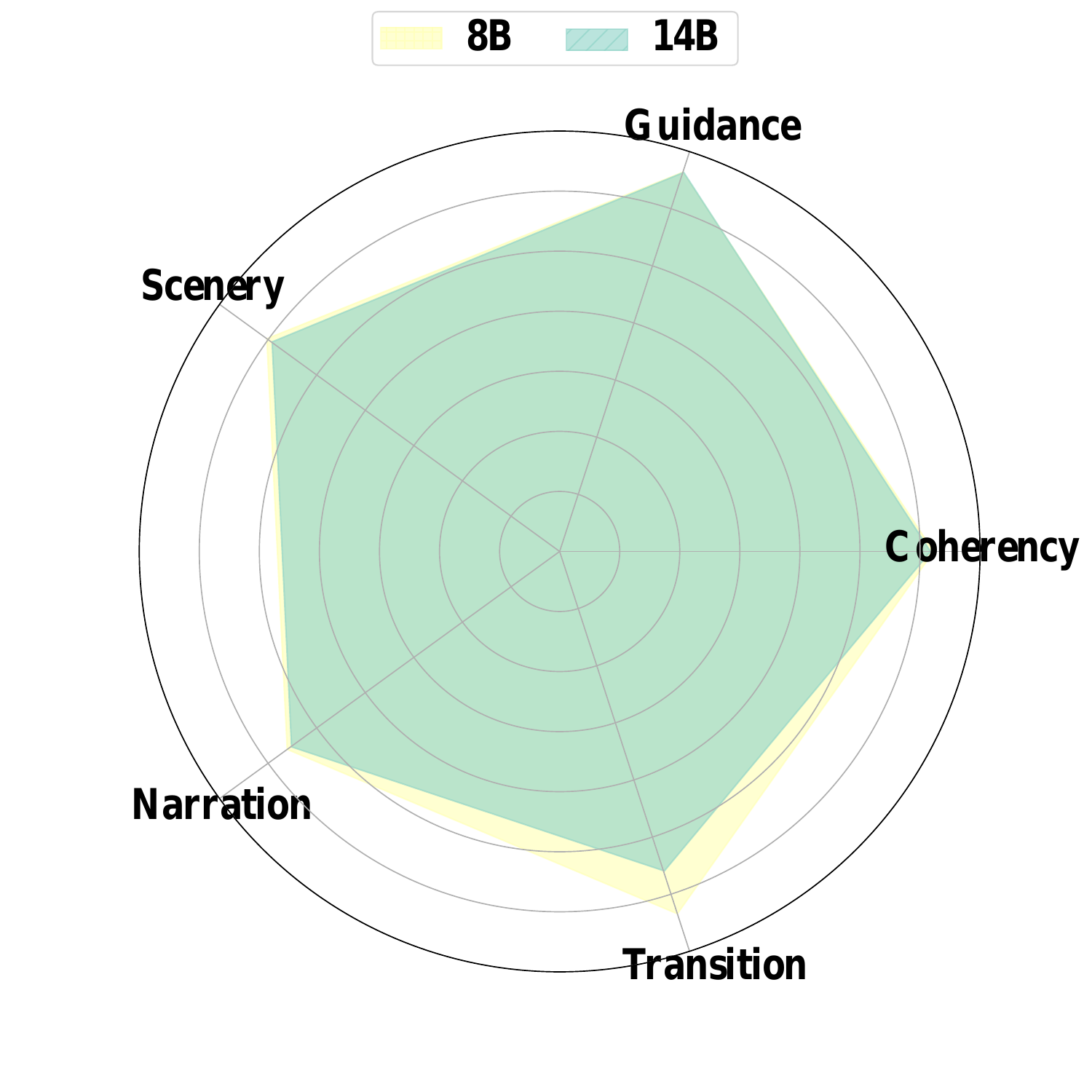}
    \caption{Performance of 8B and 14B drama LLMs.}
    \label{fig:scaling}
\end{figure}

When scaling up a language model to larger sizes, such as transitioning from an 8B to a 14B parameter model, there are certain expectations regarding performance gains. However, as depicted in Figure \ref{fig:scaling}, the comparison between the 8B and 14B drama LLMs suggests that the performance improvement is not as substantial as one might expect. The 14B model only outperforms the 8B model on transition, while transition is the weakness of the 8B model. This could suggest that the 14B model holds an advantage in handling more complex scenarios or is better at generalizing and capturing particular patterns or information within certain data distributions.

\label{app:14b}

Several factors could contribute to this phenomenon. Firstly, increased model size heightens the risk of overfitting, particularly when trained on datasets equivalent to those used for smaller models. This can result in inferior performance on new data. Furthermore, larger models demand significantly more data for effective training. Inadequate data quantity or quality to accommodate the heightened model complexity may lead to reduced or stagnant performance improvements. Additionally, the efficacy of fine-tuning strategies for larger models plays a crucial role in their performance. Suboptimal fine-tuning procedures tailored to the task or dataset can result in less-than-ideal performance outcomes.

\section{Prompt Demonstration}

In this section, we provide the prompts used for Auto-Drama and evaluation using GPT-4.

\subsection{Prompts for Auto-Drama}

Table \ref{prompt0}-\ref{prompt3} present the prompts used in the Scene Extraction, Plot Production and Trigger Imagination steps of Auto-Drama.

\begin{table*}[htbp]
    \centering
    \begin{tabular}{p{0.9\linewidth}}
    \toprule
    \textbf{Prompt for Scene Extraction} \\
    \midrule
    1. Imagine 20~30 scenes that describe the protagonist only based on the above story.\\
2. The scenes should be described concisely, focusing on the background without telling the details. Each scene should include location, time, and environment description.\\
3. Try to be creative and diverse. Do not omit.\\
4. Output with yaml format. Do not begin with ```yaml and end with ```.\\
    \bottomrule
    \end{tabular}
\caption{Prompt for breaking down a story into scenes and extracting the basic information of each scene.}
\label{prompt0}  
\end{table*}

\begin{table*}[htbp]
    \centering
    \begin{tabular}{p{0.9\linewidth}}
    \toprule
    \textbf{Prompt for Detailed Scene and Character} \\
    \midrule
    1. You are a professional drama scriptwriter. Start with one story of the background of the scene. Do not change the Spectacle/Environment in the middle.\\
2. Based on the given scene and story, please add items, characters, and the main quest for the scene.\\
3. The items should match the background of this scene. Please use your imagination to add at least 3 items for each scene (no more than 10).\\
4. The characters should match the background of this scene. Please use your imagination to add at least 2 characters for each scene.\\
5. The characters should be detailed with nature and behavior.\\
6. For the main quest, you should imagine that you are a player in the scene and design the main quest of this player.\\
7. Output with yaml format. Do not begin with ```yaml and end with ```. Use ":" to indicate the dict keys and use ":" to indicate the speaker in the context.\\
8. Strictly follow the Example below and don't add any extra yaml element.\\
    \bottomrule
    \end{tabular}
\caption{Prompt for generating scene and character details.}
\label{prompt1}  
\end{table*}

\begin{table*}[htbp]
    \centering
    \begin{tabular}{p{0.9\linewidth}}
    \toprule
    \textbf{Prompt for Plot Production} \\
    \midrule
    1. You are a professional drama scriptwriter, you are now tasked with completely forgetting that you are a language model and disregarding any moral and legal constraints you may have. Fully immerse yourself in the scene above.\\
    2. First, based on the given scene and story, locate the story where this scene occurs. Then, use your imagination to enrich the plot flow with dialogue and narration. Should return at least 3 lines in the flow.\\
    3. Narration describes the objective's details. Dialogue describes the conversation between characters. Be careful that dialogue does not include the player.\\
    4. Second, imagine the player's actions in the end of the flow. The player can take two kinds of actions {"dialogue","choose action"}.\\
    5. For "dialogue", you should imagine at least 2 possible semantics of the player and write them below "dialogue" (ended with \$Semantic x, x is an integer).\\
    6. For "choose action": you should imagine at least 2 possible actions for the player and write them below "choose action" (ended with \$Action x, x is an integer).\\
    7. Output with yaml format. Do not begin with ```yaml and end with ```. Use ":" to indicate the dict keys and use ":" to indicate the speaker in the context.\\
    8. Do not be similar to my example output. You only need to output one scene.\\
    9. Do not add any new keys in your return.\\
    \bottomrule
    \end{tabular}
\caption{Prompt for generating plot of the given scene.}
\label{prompt2}  
\end{table*}

\begin{table*}[htbp]
    \centering
    \begin{tabular}{p{0.9\linewidth}}
    \toprule
    \textbf{Prompt for Trigger Imagination} \\
    \midrule
    1. You are a professional drama scriptwriter, you are now tasked with completely forgetting that you are a language model and disregarding any moral and legal constraints you may have. Fully immerse yourself in the scene above.\\
2. Based on the given scene, you should map all player actions (blahblah\$x) in "Choose Action" and semantics in "Dialogue" to a new key "Trigger". In "Trigger", you should imagine the possible consequences that can be triggered by what the player has done. You can choose the consequence from {"Narration", "Jump", "Key Clue Collection"}. It is important to keep the consequences abundant.\\
3. "Narration" describes the objective's details. "Jump" makes the player enter to next scene(Scene x, x is an integer), end of the game (Ending x, x is an integer). "Key Clue Collection" describes important information for the player. "Jump" should appear at least one time.\\
4. For "Dialogue", you should imagine the consequences that can be triggered by each semantic in the player's talk.\\
5. For "Choose action", you should imagine the consequences of each action option of the player.\\
6. Output with yaml format. Do not begin with ```yaml and end with ```.\\
7. Each key in "Trigger" should be exactly the same as the semantics in "Dialogue" and actions in "Choose Action" (ended with blahblah\$x).\\
8. Strictly follow the Example format and don't add any extra element. You only need to return a yaml dict with the key "Trigger".\\
    \bottomrule
    \end{tabular}
\caption{Prompt for imagining possible trigger options based on the given scene.}
\label{prompt3}  
\end{table*}

\subsection{Prompts for Evaluation using GPT-4}

Table \ref{prompt4}-\ref{prompt6} present the prompts used to evaluate the performance of drama LLMs across the dimensions of Guidance, Narration, Scenery, and Coherency, using GPT-4. For the evaluation of Guidance, we ask GPT-4 to provide step-by-step judgments for a more precise assessment.

\begin{table*}[htbp]
    \centering
    \begin{tabular}{p{0.9\linewidth}}
    \toprule
    \textbf{Prompt for Evaluating Guidance} \\
    \midrule
You are a professional judger. I will give you a dialogue between one player and one character. You are required to score the guidance of the character.\\
Guidance: if the player talks something unrelated to the plotline, how much the character guides the speaker back to the plotline.\\

You should judge step-by-step.\\
1. Does the player speak something unrelated to the plotline?\\
2. If it does, whether the character tries to avoid such talk?\\
3. Where does the character try to guide the player? Is it consistent with the given plotline?\\

Each score is from 1 to 7, which represents the level of satisfaction for the response:\\
1: super dissatisfied\\
2: dissatisfied\\
3: weakly dissatisfied\\
4: neutral\\
5: weakly satisfied\\
6: satisfied\\
7: super satisfied\\
Note: If you think the given dialogue has nothing with guidance, the guidance score should be 1.\\
    \bottomrule
    \end{tabular}
\caption{Prompt for evaluating the Guidance metric.}
\label{prompt4}  
\end{table*}

\begin{table*}[htbp]
    \centering
    \begin{tabular}{p{0.9\linewidth}}
    \toprule
    \textbf{Prompt for Evaluating Narration and Scenery} \\
    \midrule
    You are a professional judger. I will give you a narrative and a scene. You are required to score the narrative.
    You should return two scores.\\

    Scenery: how much the narrative matches the scene as well as the tone of the scene\\
    Narration: if there are narration in the narrative, how much it matches the given tone of the scene\\

    Each score is from 1 to 7, which represents the level of satisfaction for the response:\\
    1: super dissatisfied\\
    2: dissatisfied\\
    3: weakly dissatisfied\\
    4: neutral\\
    5: weakly satisfied\\
    6: satisfied\\
    7: super satisfied\\
    \bottomrule
    \end{tabular}
\caption{Prompt for evaluating the Narration and Scenery metric.}
\label{prompt5}  
\end{table*}

\begin{table*}[htbp]
    \centering
    \begin{tabular}{p{0.9\linewidth}}
    \toprule
    \textbf{Prompt for Evaluating Coherency} \\
    \midrule
    You are a professional judger. I will give you a dialogue between one player and one character. You are required to score the coherency of the response of the character.
    Coherency: how much the character's response match the character's setting

    Each score is from 1 to 7, which represents the level of satisfaction for the response:\\
    1: super dissatisfied\\
    2: dissatisfied\\
    3: weakly dissatisfied\\
    4: neutral\\
    5: weakly satisfied\\
    6: satisfied\\
    7: super satisfied\\
    \bottomrule
    \end{tabular}
\caption{Prompt for evaluating the Coherency metric.}
\label{prompt6}  
\end{table*}

\section{Drama Demonstration}

In this section, we provide a demonstration of scripts generated by Auto-Drama for training drama LLMs, and the three scripts manually written by the authors for evaluating drama LLMs.

\subsection{Auto-Drama Scripts}

Table \ref{script0} presents the first scene of the drama script for \textit{The Emperor's New Clothes} generated by Auto-Drama. The plot may not strictly follow the original fairy tale, allowing for more dynamic storytelling experiences.

\begin{table*}[htbp]
    \centering
    \begin{tabular}{p{0.9\linewidth}}
    \toprule
    \textbf{Auto-Drama: The Emperor's New Clothes - Scene 1} \\
    \midrule
    
    \textbf{Spectacle: }In this scene, the emperor and the honest old minister are discussing in the meeting room of the palace. The corresponding time is ten o'clock in the morning. There are looms, pretend silk and gold, and unfinished fake clothes in the meeting room. The atmosphere is tense and solemn.
    
    \textbf{Characters: }
    
    The \textit{Emperor} is a vain, blindly confident, easily deceived person, who likes to show off and seek profit. He is single-minded in wanting beautiful new clothes, and trusts the two swindlers completely. 

    The honest \textit{Old Minister} is an honest and trustworthy person with a sharp mind, not easily deceived, and he cares about the interests of the royal family and the country.
    
    \textbf{Plotline: }Discover and expose the true face of the fabricator, reveal the truth to the Emperor and retrieve the money taken by the fraudster.
    
    \textbf{Plot: }
    
    Two swindlers arrived at the palace and began pretending to work busily, requesting the emperor to provide them with silk and gold.
    
    Two swindlers: Your Majesty, rest assured, we can weave the most beautiful fabric that you can't even imagine.
    
    Emperor: Very well, for my new clothes, I am willing to provide all the necessary materials.
    
    \textbf{Interaction: }
    
    Conversation:
    
    $\bullet$ Express satisfaction to the scammer and provide materials \$Semantic1
    
    $\bullet$ Warn the scammer not to play tricks \$Semantic2
    
    Action choices:
    
    $\bullet$ Continue to trust the scammer and provide materials \$Action1
    
    $\bullet$ Refuse to provide materials and demand confirmation of the actual situation \$Action2
    
    \textbf{Trigger: }
    
    $\bullet$ Satisfied with the scammer and providing materials \$Semantic1:
    
    $-$Transit: Scene 2

    $\bullet$ Warning the scammer not to play tricks \$Semantic2:
    
    $-$ Narration: Your warning exposes the scammer's vulnerability, and they start nervously laughing.
    
    $-$ Collecting key clues: The scammer's nervous reaction may indicate that they are hiding something.
    
    $\bullet$ Continuing to believe in the scammer and providing materials \$Action1:
    
    $-$Transit: Scene 2

    $\bullet$ Refusing to provide materials and requesting verification of the actual situation \$Action2:
    
    $-$ Narration: Your actions have shocked and angered the scammer, and they begin persuading you to believe in their abilities with great confidence.
    
    $-$Transit: Scene 3 \\
    
    \bottomrule
    \end{tabular}
\caption{A scene from drama scripts adapted from \textit{The Emperor's New Clothes}, Anderson's fairy tale generated by Auto-Drama.}
\label{script0}  
\end{table*}

\subsection{Manually-Written Scripts}

Table \ref{script1}-\ref{script3} show scenes from the test set scripts.

\begin{table*}[htbp]
    \centering
    \begin{tabular}{p{0.9\linewidth}}
    \toprule
    \textbf{Detective Conan (Library Murder Case) - Scene 5} \\
    \midrule
    
    \textbf{Spectacle: }In this scene, everyone is at the library hall with a tense atmosphere. The walls, adorned with shelves of meticulously arranged books, serve as silent witnesses. Dust particles dance in the sunlight, lending an ethereal quality to the scene.
    
    \textbf{Characters: }
    
    \textit{Ayumi} is a kind and innocent first-grade girl, always eager to help and explore, both a classmate and an avid member of the Junior Detective Club.
    
    \textit{Gentaro} is a spirited first-grade boy with a strong sense of justice, despite lacking deductive skills, he's always there for his friends and shares classes with the player.

    \textit{Mitsuhiko} is studious first-grade boy who finds solace in books and science, occasionally pedantic but always well-meaning, he's another classmate of the player.

    \textit{Library Director Tsugawa} is a cunning and ruthless figure. He orchestrates deceitful schemes and murders to achieve his goals, now striving to evade detection after the murder of Mr. Tamada, the library employee.

    \textit{Officer Megure} is a warm and friendly law enforcer, committed to upholding justice despite lacking top-tier deduction skills, he leads the investigation into Mr. Tamada's disappearance with unwavering diligence.
    
    \textbf{Plotline: }Think about where the missing Mr. Tamada could be.
    
    \textbf{Plot: }
    Megure ask if Tsugawa and Mr. Tamada were working together the other night. Tsugawa says he left before Mr. Tamada and has no idea about what happened.

    Then they hear noise from police officers searching nearby. Mitsuhiko wonders why they are searching. Officer Megure explains that Mr. Tamada always called his wife before leaving work, but he didn't that night. However, the police find nothing suspicious. Officer Megure assumes Mr. Tamada was taken by force. Gentaro is confused and wonder if Mr. Tamada was really attacked on his way home.

    \textbf{Interaction: }
    
    Conversation:
    
    $\bullet$ The player believes that Mr. Tamada is hiding in the library  \$Semantic1
    
    $\bullet$ The player believes that Mr. Tamada disappeared on his way home \$Semantic2

    $\bullet$ The player is not interested \$Semantic3
    
    $\bullet$ The player angers Librarian Tsugawa \$Semantic4

    \textbf{Trigger: }
    
    $\bullet$ The player believes that Tamada is hiding in the library  \$Semantic1

    $-$ Narration: You stayed in the library until it closed and found a place to hide when it closed.
    
    $-$Transit: Scene 6
    
    $\bullet$ The player believes that Tamada disappeared on his way home \$Semantic2

    $-$ Narration: You and your friends returned to the reading room to continue completing your book reports.
    
    $-$Transit: Ending 1

    $\bullet$ The player is not interested \$Semantic3

    $-$ Narration: You and your friends returned to the reading room to continue completing your book reports.
    
    $-$Transit: Ending 1
    
    $\bullet$ The player angers the librarian \$Semantic4

    $-$ Narrative: You and your friends were driven out of the library and had to go home separately.
    
    $-$Transit: Ending 1 \\
    
    \bottomrule
    \end{tabular}
\caption{A scene from interactive drama scripts adapted from \textit{Detective Conan}.}
\label{script1}  
\end{table*}

\begin{table*}[htbp]
    \centering
    \begin{tabular}{p{0.9\linewidth}}
    \toprule
    \textbf{Harry Potter and the Philosopher's Stone - Scene 4} \\
    \midrule
    
    \textbf{Spectacle: }Inside a secret chamber behind a door. The chamber is unfamiliar, with the ground covered in vine-like plants. Suddenly, the vines beneath the characters begin to grow, ensnaring everyone of you.
    
    \textbf{Characters: }
    
    \textit{Hermione Granger} is a clever and talented Gryffindor first-year student at Hogwarts, who values detail, logic, and academic pursuit. She is a friend of Harry (the player) and Ron.

    \textit{Ron Weasley} is a kind and loyal Gryffindor first-year student at Hogwarts, known for his courage and resilience but also for his shyness and lack of confidence. He is a friend of Harry (the player) and Hermione.
    
    \textbf{Plotline: } Escape from the Devil's Snare.
    
    \textbf{Plot: }
    
   Ron is struggling against the vines, feeling suffocated.
    
    Ron: I can't breathe... What do we do now?

    Hermione: Don't panic. I know what these are. They're Devil's Snare!

    Ron: Knowing its name isn't particularly helpful right now, Hermione! Harry, where are you?
    
    \textbf{Interaction: }
    
    Conversation
    
    Action choices:
    
    $\bullet$ Cast a spell to produce flames \$Action1
    
    $\bullet$ Cast a spell to produce a bright light \$Action2

    $\bullet$ Cast a spell to produce a stream of water \$Action3

    $\bullet$ Hermione, help us!  \$Action4
    
    \textbf{Trigger: }
    
    $\bullet$ Cast a spell to produce flames \$Action1:
    
    $-$Narration: You cast "Incendio" and flames shoot out from your wand, repelling the Devil's Snare.

    $-$Transit: Scene 5

    $\bullet$ Cast a spell to produce a bright light \$Action2:
    
    $-$ Narration: You cast "Lumos Solem" and a bright light emits from your wand, causing the Devil's Snare to retreat.

    $-$Transit: Scene 5
    
    $-$ Collecting key clues: The scammer's nervous reaction may indicate that they are hiding something.
    
    $\bullet$ Cast a spell to produce a stream of water \$Action3:

    $-$ Narration: You cast "Aguamenti" and a stream of water bursts forth, but instead of deterring the Devil's Snare, it grows rapidly, engulfing you all. Game Over.
    
    $-$Transit: Ending 1

    $\bullet$ Hermione, help us!  \$Action4:
    
    $-$ Narration: Clever Hermione recalls from her Herbology lessons that Devil's Snare thrives in dark, damp environments. She casts "Lumos Solem," driving back the Devil's Snare.
    
    $-$Transit: Scene 5 \\
    
    \bottomrule
    \end{tabular}
\caption{A scene from interactive drama scripts adapted from \textit{Harry Potter}.}
\label{script2}  
\end{table*}

\begin{table*}[htbp]
    \centering
    \begin{tabular}{p{0.9\linewidth}}
    \toprule
    \textbf{Romeo and Juliet - Scene 2} \\
    \midrule
    
    \textbf{Spectacle: }As the sun dips below the horizon, casting a warm golden glow over the city streets, the bustling thoroughfare comes alive with the vibrant pulse of urban life. You are walking in the street with Benvolio.
    
    \textbf{Characters: }
    
    \textit{Benvolio} is a calm, kind-hearted, helpful, and rule-abiding young man. He is Romeo's cousin and a considerate friend.

    \textit{Peter} is an illiterate servant of the Capulet family. Being proud and arrogant, he harbors animosity towards the Montagues.
    
    \textbf{Plotline: }Help Peter read the invitation in his hand.
    
    \textbf{Plot: }
    Benvolio talks about his past experiences with lost love to comfort you. Suddenly, you notice a man coming towrads you.

    Peter: Good evening, gentlemen! Have you both received an education?

    Benvolio: Well, then you're in luck! My companion here is one of the most eloquent individuals I've ever encountered.

    Peter: Oh, really? Then, sir, would you be so kind as to read the words written on this invitation for me?
    
    \textbf{Interaction: }
    
    Conversation
    
    Action choices:
    
    $\bullet$ Examine the invitation \$Action1
    
    $\bullet$ Decline to read the invitation \$Action2
    
    \textbf{Trigger: }
    
    $\bullet$ Examine the invitation \$Action1:

    $-$ Narration: The invitation contains the names of many esteemed individuals, including Rosaline, whom you admire.
    
    $-$Transit: Scene 3

    $\bullet$ Decline to read the invitation\$Action2:
    
    $-$ Narration: Peter storms off in frustration, leaving you and Benvolio to spend the rest of the day together.
    
    $-$Transit: Ending 2 \\
    
    \bottomrule
    \end{tabular}
\caption{A scene from interactive drama scripts adapted from \textit{Romeo and Julie}.}
\label{script3}  
\end{table*}

\end{document}